\documentclass[letterpaper]{article} 
\usepackage[]{aaai25}  
\usepackage{times}  
\usepackage{helvet}  
\usepackage{courier}  
\usepackage[hyphens]{url}  
\usepackage{graphicx} 
\usepackage{natbib}  
\usepackage{caption} 
\usepackage{algorithm}
\usepackage{algpseudocode}

\usepackage{booktabs}
\usepackage{amsfonts}
\usepackage{amsmath}
\usepackage{cleveref}
\usepackage{nicefrac}
\usepackage{xcolor}

\usepackage{newfloat}
\usepackage{listings}
\DeclareCaptionStyle{ruled}{labelfont=normalfont,labelsep=colon,strut=off} 
\floatstyle{ruled}
\newfloat{listing}{tb}{lst}{}
\floatname{listing}{Listing}
\title{QBI: Quantile-Based Bias Initialization for Efficient\\ Private Data Reconstruction in Federated Learning}
\author{
    Micha V. Nowak\textsuperscript{\rm 1},
    Tim P. Bott\textsuperscript{\rm 2},
    David Khachaturov\textsuperscript{\rm 3}, \\
    Frank Puppe\textsuperscript{\rm 1},
    Adrian Krenzer\textsuperscript{\rm 1},
    Amar Hekalo\textsuperscript{\rm 1}
}
\affiliations{
    \textsuperscript{\rm 1}University of Würzburg
    \textsuperscript{\rm 2}University of Augsburg
    \textsuperscript{\rm 3}University of Cambridge

}

\usepackage{bibentry}

\begin{document}

\maketitle

\begin{abstract}
Federated learning enables the training of machine learning models on distributed data without compromising user privacy,
as data remains on personal devices, and only model updates, such as gradients, are shared with a central coordinator.
However, recent research has demonstrated that the central entity can perfectly reconstruct private data from these
shared model updates by maliciously initializing the model's parameters. In this paper, we introduce QBI, a novel method for
malicious model initialization that significantly outperforms previous approaches in the domain of perfect user data reconstruction.
Moreover, QBI drastically simplifies the initialization process, as it does not require any training data
or knowledge about the target domain. The key insight behind QBI is the assumption that normalized input features follow a normal distribution,
which enables direct calculation of bias values that produce sparse activation patterns, thereby facilitating data
extraction through gradient sparsity.
Measured by the percentage of samples that can be perfectly reconstructed from batches of various sizes, our
approach achieves substantial improvements over previous methods, with gains of up to  \mbox{50\% on ImageNet} and up to
60\% on the IMDB sentiment analysis text dataset.
Additionally, we propose PAIRS, an algorithm that builds on QBI and can be deployed
when a separate dataset from the target domain is available, further increasing the percentage of data that can be fully
recovered. Further, we establish theoretical limits for attacks leveraging
stochastic gradient sparsity, providing a foundation for understanding the fundamental constraints of these attacks.
These limits are empirically assessed using synthetic datasets. Finally, we propose AGGP, a defensive framework
designed to prevent gradient sparsity attacks, contributing to the development of more secure and private federated
learning systems.
\end{abstract}

%
 \begin{links}
     \link{Code}{https://github.com/mvnowak/QBI}
\end{links}
\section{Introduction}
The proliferation of mobile devices and the Internet of Things has led to an unprecedented amount of data being
generated at the edge of the network~\cite{iot}. This data, often in the form of user-generated content,
sensor readings, or other types of user interactions, holds immense value for training machine learning
(ML) models~\cite{wang2021electricity}. Due to the sensitive and often private nature of this data,
traditional ML approaches that rely on centralized data collection and processing are often inadequate
from a legal or ethical perspective~\cite{basu2020ethics}. Furthermore, regulations such as data sovereignty
laws and cross-border data transfer restrictions (e.g., GDPR, CCPA) can hinder the movement of data between
jurisdictions~\cite{gdpr, ccpa}.
Federated learning (FL) was proposed as a solution to these challenges~\cite{mcmahan2017communication}.
It enables the collaborative training of ML models while preventing the need for users' data to leave
their devices -- each device computes model updates locally, which are then sent to a central entity for
aggregation into a shared model. FL, in theory, should preserve users' privacy and adhere to data
transfer restrictions, as the computed gradient updates should not expose any user data.
However, a large body of prior work has demonstrated that the FL protocol is vulnerable to multiple
forms of data reconstruction attacks, which can be roughly classified into two major categories.

\paragraph{Passive Gradient Leakage Attacks} In this scenario, the attacker is assumed to have no control
over the model's architecture, its parameters, or the specific FL protocol that is being used. The gradients
created through a standard FL protocol could either be obtained by an \textit{honest but curious} central entity,
or an external actor via a man-in-the-middle (MITM) attack. A long line of work has demonstrated, that
simply by obtaining these gradients, an attacker could learn about properties of the data~\citep{prop1, prop2},
data membership~\citep{prop2}, and even partially reconstruct user data~\citep{passive1, passive2, passive3}.

\paragraph{Malicious Model Modifications} The second major threat scenario involves ML models, whose architecture
or parameters have been deliberately modified to compromise users privacy, by increasing the amount of
data leaked through the generated gradients. This could be accomplished, either by a malicious central entity,
such as a rogue employee at a large company~\cite{boenisch2023curious}, or by an MITM attacker, intercepting the benign model and forwarding a malicious
initialization to the user~\cite{mitmattack}. A broad range of research has outlined a multitude of possible active attacks, that often leverage induced
gradient sparsity~\citep{boenisch2023curious, passive1, zhao2023secure} or sample
disaggregation~\cite{seer, pasquini2022eluding} to recover user data.

\paragraph{Efficient Perfect User Data Reconstruction with Bias Tuning}
In this work, we propose QBI, a novel method of maliciously initializing a model, that achieves significant improvements above
the previous state-of-the-art in perfect reconstruction, with gains of up to $50\%$ on ImageNet and up to $60\%$ on the IMDB
sentiment analysis text dataset. Moreover, QBI drastically simplifies the initialization process, as it removes the need
for training data, experimentally determined hyperparameters~\cite{boenisch2023curious} or handpicked target features or
classes~\cite{wen2022fishing}. This is achieved, by directly solving for bias values, which lead to sparse activation
patterns in a linear layer, therefore enabling data reconstruction via gradient sparsity.
We provide two variants of our
approach: \textit{Quantile-Based Bias Initialization} (QBI), which directly determines the optimal bias
with near-zero computational cost, and \textit{Pattern-Aware Iterative Random Search} (PAIRS) which builds on QBI,
and can be deployed if training data from the target domain is available, to further enhance reconstruction success
by incorporating auxiliary data and incurring marginally higher computational overhead.
Additionally, we derive boundaries for the expected success of attacks leveraging
stochastic gradient sparsity, which enables us to identify the inherent constraints of such attacks.
Using the findings described in this paper, we propose a novel defensive framework -- AGGP -- aimed at
preventing gradient sparsity attacks, contributing to the development of more secure and private FL systems.

\noindent \textbf{Contributions}
\begin{itemize}
    \item We establish theoretical limits for attacks leveraging stochastic gradient sparsity and empirically
    assess these limits using synthetic datasets.
    
    \item We propose a novel, compute efficient method of maliciously initializing fully-connected layers in an active
    attack on the FL protocol, which achieves state-of-the-art results in the domain of perfect user
    data reconstruction, while simplifying the initialization process, by removing the need for training data or
    handpicked target features or classes.
    
    \item We provide two variants of our approach: QBI which can be deployed with near-zero computational cost,
    and PAIRS which can be used to achieve increased performance if auxiliary data and increased compute is available.
    We release an open-source implementation of our method, including dedicated
    scripts to reproduce all results reported in this paper.
    
    \item We extensively evaluate both QBI and PAIRS and find that we achieve improvements above the previous
    state-of-the-art in perfect reconstruction with gains of up to $50\%$ on ImageNet and up to $60\%$ on the IMDB
    sentiment analysis text dataset.

    \item We propose and evaluate AGGP, a novel and compute efficient \textit{defensive framework} that prevents data
    leakage from both active and passive attacks that leverage gradient sparsity in fully connected layers.
\end{itemize}

\section{Background}
\paragraph{Federated Learning}
FL is a decentralized approach to machine learning that enables multiple parties to collaboratively train a shared
model on their local data without sharing the data itself. Each client has a local dataset and computes an update to
the model parameters based on their local data. The update is typically computed as the gradient of the local loss
function with respect to the model parameters. The local updates are then aggregated to update the global model
parameters. One common aggregation method is federated averaging, which computes the weighted average of the local
updates. This process is repeated over multiple rounds to train the global model~\cite{mcmahan2017communication}.

\paragraph{Gradient Sparsity Attacks} 
As demonstrated by~\citet{geiping2020inverting}, it is feasible to extract a single input from the gradients of a
fully-connected layer, given that the layer is preceded only by fully connected layers, has a bias $b$, uses the
ReLU activation function, and the gradient of the loss with respect to the layer's output contains at least one
non-zero entry (see Proposition D.1 in~\citet{geiping2020inverting} for a full proof). If this layer is placed
at the beginning of the network, this corresponds to reconstructing the original input data point $x$. As shown
in Section 5.1 of~\citet{boenisch2023curious}, for any non-zero output $y_i$, the gradient of the corresponding
weight row directly contains the input scaled by the gradient of the loss with respect to the bias. In practice,
gradients are typically computed as the average over an entire batch of samples. Since a single neuron is often
activated by multiple samples, the gradients of its weight row contain the average of multiple input data points,
effectively obscuring them and preventing individual extraction. To extract an input sample $x$, there has to
exist a neuron $n_i$ such that $L(x)_i > 0$ and $L(x')_i < 0$ for all $x' \neq x$, where $L(x)_i$ denotes the
activation of the $i$-th neuron for sample $x$. Since the samples that are to be extracted are unknown, initializing
a layer to achieve this specific activation pattern is challenging. Therefore, the goal of stochastic gradient
sparsity attacks is to increase the probability that a neuron is activated only by a single sample, while producing
a variety of neurons with diverse activation patterns, to capture as much data from the target domain as possible.

\paragraph{Adaptability to CNN-Based Architectures}
\label{sec:cnn_adaptability}
\citet{boenisch2023curious} extend the gradient sparsity attack to Convolutional Neural Networks (CNNs), where one
or more convolutional layers precede the first linear layer. By utilizing zero-padding, a stride of one, and
maliciously initialized filters, convolutional layers can effectively be transformed into identity functions,
which perfectly transmit the inputs deeper into the network, allowing them to be extracted once they pass through
a fully connected layer. For a detailed description, including visualizations, see Appendix B of~\citet{boenisch2023curious}.

\section{Related Work}
In FL, a common threat model is the passive gradient leakage scenario, where a malicious entity obtains a set
of gradients to recover the original data points. Previous research has demonstrated that these gradients can
be used to infer data properties~\citep{prop1, prop2} or data membership~\citep{prop2}. Several optimization-based
attacks have been proposed~\citep{yin2021see, passive3, zhu2019deep}, which optimize a batch of random noise to
generate gradients similar to those observed and thereby achieve partial reconstruction of user data. Although
applicable to various architectures, these methods often require significant computational resources and are unable
to achieve perfect reconstructions.

The second threat model, encompassing so-called active attacks, assumes that the attacker can not only obtain the
gradients but also maliciously modify the model's architecture or parameters before it is shared with the client. This
could be accomplished either by a malicious server (MS) or through a man-in-the-middle (MITM) attack~\cite{mitmattack}.
The objective of these malicious model modifications is to enable the reconstruction of a larger amount of private data
from the gradients generated by the clients. Research in this area has identified various active
attacks that exploit induced gradient sparsity~\citep{boenisch2023curious, passive1, zhao2023secure} or sample
disaggregation techniques~\citep{seer, pasquini2022eluding} to recover user data. The SEER framework~\citep{seer}
is a notable example of an MS attack that avoids client-side detection in gradient space by disaggregating samples
in an embedding space that is unknown to the client. Although it achieves a high percentage of well-reconstructed
images, it incurs a high computational cost (14 GPU days to train on an A100 with 80GB) and falls short of
perfectly reconstructing data.

We mainly build on the work of~\citet{boenisch2023curious} that introduced the concept of \textit{trap weights}, a
computationally efficient way to initialize a model to induce gradient sparsity. By adding a slight negative shift
to the weight values, their approach achieves perfect reconstruction on multiple datasets. However, their method
relies on an experimentally determined scaling factor. In contrast, we automate the model's initialization process
and achieve substantially higher rates of perfect reconstruction.

\section{Method: Adversarial Bias Tuning}
\label{section:AdvBiasTuning}
\paragraph{The Threat Model}
Our method assumes an active attacker that is capable of modifying an ML model's parameters, and obtaining
the subsequently produced gradients. The attackers objective is to increase the amount of private user data
which can be reconstructed from these gradients. This could either be accomplished by a malicious server (MS) during
the initialization of the protocol, or by a man-in-the-middle (MITM) attacker, intercepting the communication between a
benign server and its client, modifying the model parameters before forwarding the model to the client.
For the attack to be possible, the model must include a linear layer with a ReLU activation function, either at the beginning
of the network, or positioned such that it is preceded only by \textit{identity-capable} layers, i.e., by layers whose
parameters can be modified to effectively turn them into identity functions. This could include convolutional layers (see ~\Cref{sec:cnn_adaptability}), additional linear layers with weight matrices set to the identity matrix, or any other layers wrapped by residual connections, where their parameters can be set to zero, effectively making their output equivalent to an identity function when combined with the residual connection.
For our primary attack method QBI, the attacker
does not require any training data or knowledge about the data in the target domain.
Our second method, PAIRS, can be applied in scenarios where the attacker does have access to a dataset of samples
from the target domain. These datapoints can be acquired in several ways. Firstly,
the attacker might use a public dataset from the target domain, such as a dataset of medical images. If no such
dataset is available, the attacker could conduct a passive attack by collecting samples leaked from the
gradients generated by a benign ML model. Alternatively, the attacker could use samples obtained from previous active
attacks on the same or different clients, where QBI was employed. The number of required datapoints to employ PAIRS
equals the size $N$ of the linear layer that the attacker aims to maliciously modify.

\paragraph{Intuition of Bias Tuning} This section summarizes the key insights
of our statistical approach, for the complete derivations refer to~\Cref{sec:biastune_derivations}.
Given a linear layer $L$ of shape $N \times M$, that uses the ReLU activation function,
and a batch $\tilde{X}$ of $B$ samples $x_1, \cdots, x_B$, the objective is that for every $x$ there exists one and only
one neuron $n_i$ such that $L(x)_i > 0$ and $L(x')_i < 0$ for all $x' \neq x$, where $L(x)_i$ denotes the activation of
the $i$-th neuron for sample $x$. Such a neuron $n_i$ allows for perfect reconstruction of $x$ from the
gradients of its weight row, while producing zero gradients for all other samples. Therefore, for every
neuron $n_i$ the desired probability to activate for any sample in the batch is $1/B$. We take the model's weights
and our approximated normalized features to be independently and identically distributed (i.i.d.) random variables
sampled from a normal distribution:
\begin{equation}    
\begin{pmatrix}
\bf{w}_i\\ 
\bf{x}_i
\end{pmatrix}\sim\mathcal{N}(0,I_{2M})
\end{equation}
Using this assumption, the probability of a single neuron, with corresponding weight row $w_i$ and bias $b_i$,
activating for a sample $x_i \in \mathbb{R}^{M}$ can be expressed as:
\begin{equation}
    P\bigl(L(x_i)>0\bigr) = P\bigl(w_i^{T}x_i+b_i > 0 \bigr),
\end{equation}
which can be expanded as:
\begin{equation}\label{eq:summErweitert}
\begin{split}  
     P\bigl(w_{i1}x_{i1}+\ldots +w_{iM}x_{iM}+b_i > 0 \bigr)
\end{split}
\end{equation}
As a sum of i.i.d. random variables with existing second moments,
we can make a statement about convergence in distribution:
\begin{equation}
P\bigl(w_{i1}x_{i1}+\ldots +w_{iM}x_{iM}+b_i > 0 \bigr)\xrightarrow{d}\Phi\biggl(\frac{b_i}{\sqrt{M}}\biggr)
\end{equation}
where $\Phi$ is the cumulative distribution function of a standard normal distribution.
Meaning we can solve for the asymptotically optimal bias $b_*$ by using the inverse cdf or quantile function:\begin{equation}\label{eq:b*}
    b_*=\Phi^{-1}(\frac{1}{B})\cdot \sqrt{M}
\end{equation}

\paragraph{Extraction metrics} We closely follow the extraction metrics proposed by~\citet{boenisch2023curious}.
Given a linear layer with $N$ output neurons and a batch $\tilde{X}$ of $B$ samples $x_1, \cdots, x_B$, we define
the following metrics:
\begin{enumerate}
     \item \textbf{Active neurons} ($A$): This metric represents the percentage of neurons in
     layer $L$ that activate for at least one of the $B$ samples. Formally, let $N_A$ be the
     number of neurons $n_i$ that satisfy the condition that there exists at least one
     input $x$ in $\tilde{X}$ such that $L(x)_i > 0$, where $L(x)_i$ denotes the activation
     of the $i$-th neuron in $L$ for sample $x$.The active neurons metric $A$ is therefore
     defined as the ratio of $N_A$ to the total number of neurons $N$:
    \begin{equation}
    A = \frac{N_A}{N}
    \label{eq:defA}
    \end{equation}
  \item \textbf{Extraction-Precision} ($P$): This metric measures the percentage of neurons that allow for
     the extraction of individual data points. Specifically, let $N_u$ be the number of
     neurons $n_i$ in layer $L$ with unique activations, i.e. those that satisfy the following
     condition: $L(x)_i > 0$ for one input $x$ in $\tilde{X}$, and $L(x')_i < 0$ for all other
     inputs $x' \neq x$. The extraction-precision $P$ is defined
     as the following ratio:
    \begin{equation}
    P = \frac{N_u}{N}
    \label{eq:defP}
    \end{equation}
    \item \textbf{Extraction-Recall} ($R$): The extraction-recall measures the percentage of input
     data points that can be perfectly reconstructed from any gradient row. Let $B_0$ be the
     number of data points that can be extracted with an $l_2$-error of zero, then $R$ is denoted as:
    \begin{equation}\label{eq:Rdef}
        R=\frac{B_0}{B}
    \end{equation}
\end{enumerate}
Notably, $R$ is the most significant metric, as neither $A$ nor $P$ can be used in isolation to
meaningfully assess the effectiveness of an attack. A high $A$ value could lead to overlapping
activations that prevent individual extraction, while a high $P$ value could be observed in a
scenario where all neurons activated for the same sample. If the true probability of a neuron
activating is $1/B$, we can derive the explicit probabilities $p_{A;B}$ and $p_{u;B}$ for a
neuron to be counted as a success in the context of the $A$ and $P$ metrics, respectively.
Since the success of one neuron does not influence the remaining neurons, the entire batch
follows a binomial distribution: $N_A\sim\mathcal{B}(N,p_{A;B})$ and $N_u\sim\mathcal{B}(N,p_{u;B})$.
Consequently, the expected activation share $A$ and the expected extraction-precision $P$ are:
\begin{equation}
    p_{A;B}=\mathbb{E}[A_B]=1-\bigl(\frac{B-1}{B}\bigr)^B
    \label{eq:expA}
\end{equation}
and
\begin{equation}
    p_{u;B}=\mathbb{E}[P_B]=\bigl(\frac{B-1}{B}\bigr)^{B-1}
    \label{eq:expP}
\end{equation}

For growing batch sizes, these converge to:
 \begin{equation}
     \lim\limits_{B \to \infty} \mathbb{E}[A_B] = 1-\frac{1}{e}\approx 63.2\%
     \label{eq:optimalA}
 \end{equation}
and
  \begin{equation}
     \lim\limits_{B \to \infty} \mathbb{E}[P_B] = \frac{1}{e}\approx 36.8\%.
     \label{eq:optimalP}
 \end{equation}
Assuming the perfect activation probability of $1/B$ has been achieved, the expected share of data points that the
malicious actor can perfectly reconstruct is:
\begin{equation}
   p_{R;B;N} = \mathbb{E}[R] = 1-\left(1-\frac{1}{B}\left(\frac{B-1}{B}\right)^{B-1}\right)^{N}
   \label{eq:R}
\end{equation} 
See~\Cref{sec:biastune_derivations} for a more detailed explanation. As~\Cref{eq:R} assumes the optimal scenario, only
obtainable with per-neuron activation probabilities of $1/B$, it provides an upper limit for the expected success of
stochastic gradient sparsity attacks on real-world data, which will yield lower expected results, the further the
normalized data deviates from being normally distributed. In~\Cref{table:synthetic} we assess these boundaries by
evaluating the extraction success on synthetic truly random datasets.

\paragraph{Quantile-Based Bias Initialization (QBI)} QBI is the primary active attack method on the FL protocol that we propose. It maliciously modifies the parameters of a linear layer $L$ before sending the modified model to a client $k$ targeted for extraction. The weight values $w$ of $L$ are initialized from a
standard normal distribution. Given a batch size $B$ used on the client side and the number of input
features $M$, QBI determines the bias value $b^*$ using~\Cref{eq:b*}, which approximately leads to an
activation probability of $1/B$ for each neuron in $L$. Even though the true distribution of features
on the user side is unknown and the features are neither independent nor truly normally-distributed,
our approximation is effective in practice, when regular data normalization is applied.
\begin{algorithm}[t]
\caption{Pattern-Aware Iterative Random Search (PAIRS)}
\label{algo:PAIRS}
\begin{algorithmic}[1]
\State \textbf{Input:} Linear layer $L$ of shape $M \times N$, Number of retries $T$
\State $K$ Batches $\tilde{X}_{1}, \cdots, \tilde{X}_{K}$ of size $B \gets \lceil N/K \rceil$
\State \textbf{Initialize:} $L.bias \gets \phi^{-1}(\frac{1}{B})\cdot \sqrt{M}$ \Comment{Fill bias values using quantile function}

\ForAll{$\tilde{X}_k$}
    \State \textbf{Initialize:} $F_D \gets \emptyset$ \Comment Frozen data points
    \For{neuron $n = (k-1)B$ to $kB - 1$}    
        
        \For{$t = 1$ to $T$} \Comment{Random resets for $n$}
            \State $A \gets L(\tilde{X}_k)_n$ \Comment Get activations for  $n$
            \State $I \gets \{i \,|\, A[i] > 0\}$ \Comment Get active indices
            \State $s \gets I[0]$
            \If{$|I|\neq1 \vee s \in F_D$} \Comment{Check isolation}
                        \State $L.W_i \sim \mathcal{N}$ \Comment{Re-initialize weight row $i$} 
                        \State \textbf{continue}
            \EndIf
            \State $F_D \gets F_D \cup \left \{  s\right \}$ \Comment{Mark sample as frozen}
        \EndFor
    \EndFor
\EndFor
\end{algorithmic}
\end{algorithm}
\paragraph{Pattern-Aware Iterative Random Search (PAIRS)}
We propose the PAIRS algorithm that further adapts the malicious QBI linear layer to the target domain
when auxiliary data from the target domain is available. By acknowledging that real-world data, such as images, rarely exhibits
the assumed i.i.d. properties, PAIRS iteratively searches the weight space to better capture the
underlying patterns. The procedure, outlined in~\Cref{algo:PAIRS}, begins by performing a forward
pass with a batch of auxiliary data. It then identifies and re-initializes the weight rows of neurons
that are either overactive or underactive, as well as those that exhibit redundant activation patterns.
Through this process, PAIRS builds neuron-sample pairs, iteratively searching the weight space until all
samples are covered or a fixed number of iterations is reached. Specifically, with an output shape of
$N$ and a batch size of $B$, PAIRS uses $N/B$ batches of $B$ samples for groups of $B$ neurons each.
By randomly re-initializing weight values, PAIRS avoids detectability in weight space while increasing
the percentage of data that can be perfectly reconstructed, surpassing the performance of plain QBI.

\section{Defence: Activation-Based Greedy Gradient Pruning (AGGP)}
\label{section:AGGP}
To counter attacks that exploit gradient sparsity in fully connected layers, we propose Activation-Based Greedy Gradient
Pruning (AGGP). This is a novel approach that detects and mitigates both passive and active data leakage. Unlike previous
works that suggest skipping entire training rounds when potential data leakage is detected~\citep{seer}, we adopt a more
targeted strategy by selectively pruning gradients of suspect neurons, scaled by their activation pattern. This approach accounts
for the fact that even benign networks may occasionally leak data points, and that skipping entire updates could
withhold valuable training information.

A forward hook is registered at a potentially vulnerable linear layer $L$. The hook records and caches activation
counts, i.e., the number of samples in the batch that lead to a positive activation in a particular neuron. As
outlined in~\Cref{algo:AGGP}, after the loss and gradients have been calculated, AGGP iterates over all neurons
in $L$. We take $a_n$ to be the number of samples that activate a specific neuron $n$. Take $c$ to be an arbitrary
cut-off sample count. Neurons that did not activate (i.e., with $a_n=0$) or those that activated for more than $c$
samples (i.e., with $a_n\geq c$) are skipped. For all other neurons with $0 < a_n < c$, the percentage $p_{keep,n}$
of gradient values to retain is calculated as:
\begin{equation}    
\label{eq:pkeep}
p_{keep,n} = \frac{(a_n-1)^{2}\cdot\left(p_{u}-p_{l}\right)}{(c-2)^{2}}+p_{l}
\end{equation}
where $p_l$ and $p_u$ represent the lower and upper bound for the percentage of gradient values that are
retained for $a=1$ and $a=c-1$ respectively. In our experiments on the ImageNet dataset, we set $c=16$,
$p_l=0.01$ and $p_u=0.95$. See \Cref{app:aggphyperparams} for a detailed explanation of how we arrived at
these hyperparameters. For higher $a_n$ values $p_{keep,n}$ increases as the overlapping samples lead to more
diffuse representations, where individual samples are increasingly obscured. AGGP then continues by sorting the
gradients of the corresponding weight row by their absolute magnitude. Among the top $p_{keep,n}$ percent of values,
$25\%$ are randomly selected to be retained, while all other values are set to zero, to further obscure potentially
connected features. This effectively prevents the perfect reconstruction of individual samples, while $25\%$ of the
gradient values corresponding to the $p_{keep,n}$ percent largest magnitudes are maintained, allowing the propagation
of valuable training information.

\begin{algorithm}[t]
\caption{Activation-based Greedy Gradient Pruning (AGGP)}
\label{algo:AGGP}
\begin{algorithmic}[1]
\State \textbf{Input:} Linear layer $L$ of shape $M \times N$, Batch $\tilde{X}$ of size $B$, cut-off threshold $c$,
lower and upper pruning bounds $p_l$ and $p_u$
\State $\mathbf{A} \gets L(\tilde{X}) > 0$ \Comment{Cache activations during fwd. pass}
\State $loss = \cdots$
\State $\textbf{G} \gets loss.backward()$ \Comment{Compute gradients}
\ForAll{neuron $n$ in $L$}
    \State $a_n \gets \sum_{i=1}^{M} \mathbb{I}(A_{n,i} > 0)
$ \Comment{Get activation count}
    \If{$a_n = 0 \vee a_n > c$} 
        \State \textbf{continue} \Comment{Skip sufficiently active neurons}
    \EndIf
    \State $\mathbf{p}_{keep} \gets \frac{(a_n-1)^{2}\cdot\left(p_{u}-p_{l}\right)}{(c-2)^{2}}+p_{l}$ 

    \State $k \gets \lfloor \mathbf{p}_{keep}\cdot N \rfloor$ 
    
    \State $\mathbf{s} \gets \text{argsort}(\mathbf{|G_n|)}$ 

    \State $\mathbf{I} \gets \mathbf{s}[:k]$ \Comment Greedy prune selection

    \State $\mathbf{I} \gets \mathbf{I} \cup \mathbf{s}[k:][\text{randperm}(N-k)[:\lfloor 0.75 \cdot (N-k) \rfloor]]$

    \State $\mathbf{G_n}[\mathbf{I}] \gets 0$ \Comment{Zero out gradients}

\EndFor
\end{algorithmic}
\end{algorithm}

\section{Experimental Evaluation}
\label{section:expEval}

We closely follow the experimental setup of \citet{boenisch2023curious} for both image and text data,
to allow a direct comparison to be made. See~\Cref{appendix:imp_details} for further implementation details.

\paragraph{Image Data Extraction}
We evaluated our method on two benchmark vision datasets: ImageNet~\citep{imagenet} at a resolution of $224\times224$,
and CIFAR-10~\citep{cifar} at a resolution of $32\times32$. As our method requires normalized data,
we used publically available normalization parameters for both datasets. The model we used consisted of
convolutional layers maliciously initialized to transfer the input further into the network, followed by
a linear layer initialized with either QBI or PAIRS. The rate of perfectly reconstructed images
$R$ (see~\Cref{eq:Rdef}) was evaluated using batch sizes of 20, 50, 100, and 200, and layer sizes of 200, 500, and 1000.
Our results, presented in~\Cref{table:imageResults,table:imageResults2}, surpass those reported by~\citet{boenisch2023curious}
using their \textit{trap weights} approach. Our method yields consistent improvements for both
the ImageNet and CIFAR-10 datasets across various layer size and batch size combinations. The Wilcoxon signed-rank test confirms the significance of these improvements, with p-values of $p<0.0005$ for both QBI and PAIRS when compared to the baseline method (\textit{TWs}). Notably, the most significant gains are observed on ImageNet with smaller batch sizes, where our method
achieves up to $50\%$ higher reconstruction rates. \Cref{fig:user_data} in \Cref{appendixResults} presents an
example batch of 20 images and the perfectly reconstructed subset of 16 images, obtained from a layer size of 200.
We also conducted experiments using unnormalized data, inserting either a batch normalization layer or a
layer normalization layer before the maliciously initialized linear layer. See \Cref{app:batchnorm} and \Cref{table:norms} for more details.
 
\paragraph{Text Data Extraction}
The IMDB sentiment analysis dataset~\cite{imdb} was used to evaluate the extraction of text data.
Our model operates on 250-token sentences with an embedding dimension of 250, where the embedded tokens
are directly fed to a fully connected layer of size 1000. We re-trained the \texttt{bert-base-uncased}
tokenizer~\citep{bert} on the IMDB dataset, resulting in a vocabulary size of $10,000$. The extraction
was evaluated on batch sizes of 20, 50, 100, and 200. The results, presented in~\Cref{table:textResults},
are compared to those reported by~\citet{boenisch2023curious}. Our approach performs similarly across batch
sizes 20 and 50, but achieves performance gains of $25\%$ and $47\%$ on batch sizes 100 and 200, respectively.

\paragraph{Secondary Metrics}
The advantage of our method can be explained by examining the secondary metrics precision $P$ and
activation value $A$. As established in~\Cref{eq:optimalA,eq:optimalP}, the theoretical optimum
for these values lies at $A = 1 - 1/e \approx 63.2\%$ and $P = 1/e \approx 36.8\%$. Although these can only be achieved
if the target data is truly normally distributed (see~\Cref{table:synthetic} in the Appendix),
values that lie closer to this optimum will lead to better reconstruction rates. \Cref{table:APImage} in
the Appendix compares the $A$ and $P$ values achieved using QBI and PAIRS to those obtained
using \textit{trap weights}~\cite{boenisch2023curious} on image data. Specifically, on the ImageNet dataset,
our $A$ values range from $72\%$ to $87\%$, whereas those reported in~\citet{boenisch2023curious} show
a stronger variance across batch sizes, ranging from $9\%$ to $89\%$. Similarly, our precision values
range from $26\%$ to $34\%$ on ImageNet, while those reported in~\cite{boenisch2023curious} vary from $23\%$ to $94\%$.

\begin{table}[t]
    \centering
    \caption{Comparing the percentage of perfectly reconstructed images ($R$, see~\Cref{eq:R}) taken from
    the \mbox{ImageNet} dataset, with the results reported by \citet{boenisch2023curious} using
    their \textit{trap weights (TWs)} approach, across various combinations of neuron counts $N$ and batch sizes $B$.
    The listed values represent the mean across 10 random initializations, with each initialization
    evaluated on 10 random batches. The error margins indicate the 95\% confidence interval.}\label{table:imageResults}
\begin{tabular}{@{}cccc@{}}
\toprule
\multicolumn{4}{c}{\textbf{ImageNet}}                                                  \\ 
\textbf{(N, B)} & \textit{TWs} & \textit{QBI (Ours)} & \textit{PAIRS (Ours)}    \\ \midrule
(200, 20)     & 35.5         & 82.5 $\pm$ 2.43     & \textbf{85.5 $\pm$ 1.25} \\
(200, 50)     & 30.4         & 52.0 $\pm$ 1.46     & \textbf{56.0 $\pm$ 1.13} \\
(200, 100)    & 24.0         & 29.0 $\pm$ 0.92     & \textbf{34.6 $\pm$ 0.67} \\
(200, 200)    & 11.3         & 15.1 $\pm$ 0.62     & \textbf{19.5 $\pm$ 0.60} \\
(500, 20)     & 49.0         & 93.6 $\pm$ 1.10     & \textbf{94.5 $\pm$ 0.84} \\
(500, 50)     & 42.6         & 73.5 $\pm$ 1.50     & \textbf{76.8 $\pm$ 1.27} \\
(500, 100)    & 35.8         & 49.0 $\pm$ 1.09     & \textbf{55.8 $\pm$ 0.87} \\
(500, 200)    & 19.9         & 29.2 $\pm$ 0.49     & \textbf{35.9 $\pm$ 0.35} \\
(1000, 20)    & 59.5         & 96.6 $\pm$ 0.78     & \textbf{96.7 $\pm$ 0.66} \\
(1000, 50)    & 51.6         & 84.3 $\pm$ 0.59     & \textbf{86.6 $\pm$ 0.67} \\
(1000, 100)   & 45.7         & 64.8 $\pm$ 0.57     & \textbf{68.8 $\pm$ 1.00} \\
(1000, 200)   & 28.8         & 42.7 $\pm$ 0.72     & \textbf{49.4 $\pm$ 3.00} \\ \bottomrule
\end{tabular}
\end{table}

\begin{table}[t]
    \centering
    \caption{Comparing the percentage of perfectly reconstructed text samples $R$ from the IMDB
    dataset \cite{imdb}, with the results reported by \citet{boenisch2023curious}, using their \textit{trap weights (TWs)} approach, across various
    batch sizes $B$ using a layer size of 1000. The values represent the mean across 10 random
    initializations, with each initialization evaluated on 10 random batches. The error margins
    indicate the 95\% confidence interval.}\label{table:textResults}
    \begin{tabular}{@{}cccc@{}}
        \toprule
        \multicolumn{4}{c}{\textbf{IMDB Text}}                                                  \\ 
        \textbf{B} & \textit{TWs} & \textit{QBI (Ours)}               & \textit{PAIRS (Ours)}             \\ \midrule
        20         & 100.0     & 100 $\pm$ 0.00 & 99.9 $\pm$ 0.20           \\
        50         & 96.2     & \textbf{98.9 $\pm$ 0.46} & 98.6 $\pm$ 0.43          \\
        100        & 65.4     & 90.5 $\pm$ 0.33          & \textbf{90.8 $\pm$ 0.82} \\
        200        & 25.5     & 72.8 $\pm$ 0.64          & \textbf{73.3 $\pm$ 0.78} \\ \bottomrule
    \end{tabular}
\end{table}

\begin{figure*}[t]
    \centering
    \includegraphics[width=\textwidth]{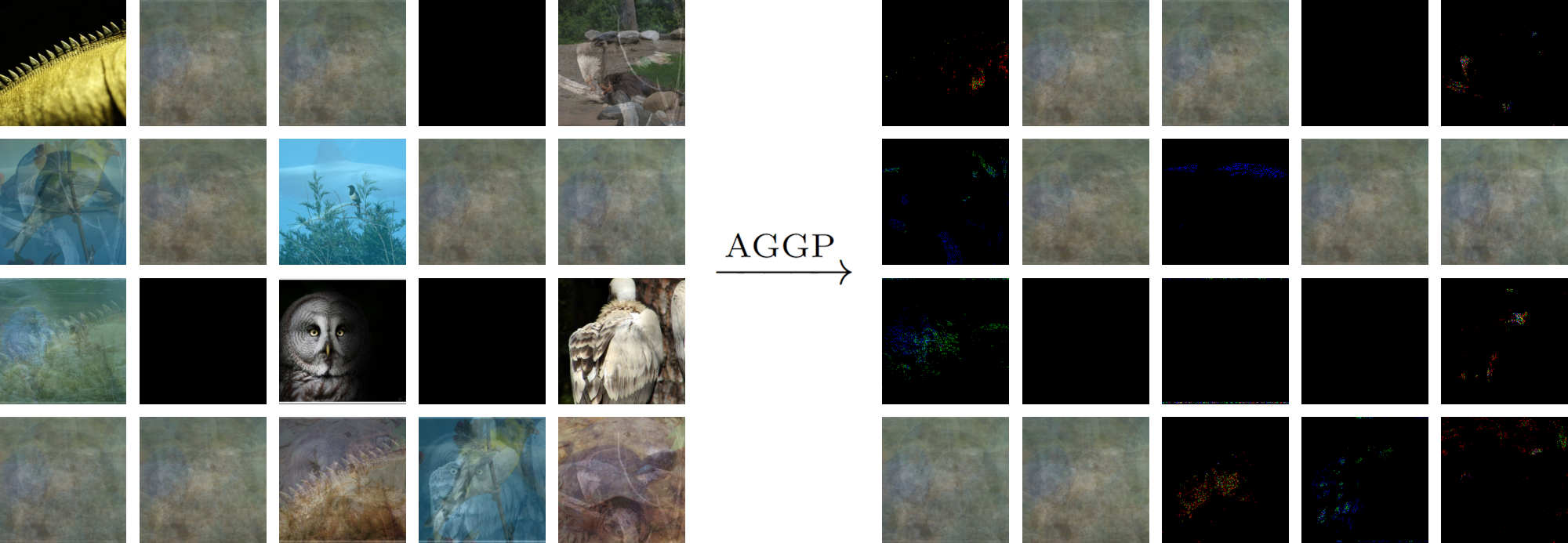}
     \caption{Visualization of the passive data leakage of the first 20 neurons of a linear layer
     of size 200 (left) and the impact of  our proposed client-side defense framework AGGP (right). Sparsely activated neurons are
     aggressively pruned, while the gradients of neurons with activation counts exceeding the
     cut-off threshold remain unaffected.}
    \label{fig:passive_leak}
\end{figure*}

\paragraph{AGGP} We find that our proposed defense framework reduces the percentage of perfectly
reconstructable samples obtained via gradient sparsity in linear layers to \textbf{zero}.
This occurs as any neuron that meets the condition for perfect extraction ($a_n=1$), triggers
the pruning of $1-0.25\cdot p_l$ percent of gradient values of its corresponding weight
row (see~\Cref{eq:pkeep}). The effect of our defense is best presented visually -- the left-hand
side of~\Cref{fig:passive_leak}, displays the data that is leaked passively from the first 20
neurons of a benign network after a single training step with a batch of 20 samples from the
ImageNet dataset. The right-hand side depicts the impact of AGGP on the same scenario, where gradients
of neurons with low activation counts are aggressively pruned, while those with high activation
counts remain unaffected. Further visualizations of AGGP's impact on a maliciously initialized
model, along with preliminary observations of its effect on training performance, are provided in~\Cref{appendix:aggp}.

\section{Discussion}
\label{section:discussion}

\paragraph{Performance under Secure Aggregation and Distributed Differential Privacy}
Secure aggregation (SA)~\cite{secureagg} and distributed differential privacy (DDP)~\cite{difprivacy} are proposed
modifications to the traditional FL protocol, designed to decrease the amount of trust users have to place in the
central entity. SA employs a multiparty computation protocol to perform decentralized gradient aggregation, while
in DDP users locally add a small amount of noise to their gradient updates. \citet{cah2} demonstrated that attacks
leveraging gradient sparsity can undermine these protocols if the server is able to introduce malicious nodes in
the form of so-called sybil devices. Since their work uses the previously described \textit{trap weights}
\cite{boenisch2023curious}, replacing the model initialization with our QBI or PAIRS approach would significantly
improve the performance of this attack vector. Our work employs the exact same mechanism (gradient sparsity) to achieve
reconstruction, which is why the findings of the referenced work directly apply to our method.
Our method introduces a novel way of inducing gradient sparsity, but the extraction mechanism remains the same;
therefore, previous results are still applicable.

\paragraph{Detectability}
Our method leaves all weight values completely randomly initialized, making it virtually undetectable in weight
space. However, it introduces a negative shift in the bias values, which could potentially be detected by the client.
Additionally, as our method relies on gradient sparsity, it would be feasible to detect it in gradient space, e.g.,
by leveraging measures like the disaggregation signal-to-noise ratio~\cite{seer}.

\paragraph{Limitations}
Dimensionality reducing or lossy layers, such as MaxPooling or Dropout, which precede the maliciously initialized linear layer, will
diminish the fidelity and thereby prevent perfect data extraction. We conduct preliminary tests on the impact of AGGP on
training performance that indicate little-to-no adverse effects, however, a comprehensive evaluation across a wider
range of architectures, datasets, and pruning functions is needed to achieve generalizable insights.

\section{Conclusion}
\label{section:conclusion}
In this work, we introduce a novel malicious model modification method, designed to enhance attacks targeting private data reconstruction
in federated learning systems. Our primary attack method, QBI, outperforms existing gradient sparsity methods across various datasets and batch sizes, achieving superior rates of perfect user data reconstruction. Moreover, QBI simplifies the initialization process by eliminating the need for training data or prior knowledge about the target domain. We also introduce PAIRS, a more sophisticated attack method that can be deployed if data from the target domain is available, to achieve further performance gains.
By establishing theoretical limits for stochastic gradient sparsity attacks, our work provides a crucial step towards
a more comprehensive understanding of the fundamental constraints posed by the probabilistic nature of these attacks.
Additionally, we introduce AGGP as a defense mechanism against gradient sparsity attacks, effectively mitigating data
leakage in linear layers. Although our attack method poses privacy risks, we believe that sharing the details of our approach will encourage systematic exploration of privacy safeguards and enable practitioners to better assess
and mitigate privacy risks in FL deployments, ultimately promoting safer machine learning practices.

\bibliography{aaai25}
\newpage
\appendix

\section{Background}
\label{sec:biastune_derivations}
This section represents an expanded version of the theoretical foundations we introduce in~\Cref{section:AdvBiasTuning},
including intermediate derivations and an additional analysis from a graph theory perspective. Given a linear layer
$L$ of shape $N \times M$, that uses the ReLU activation function,
and a batch $\tilde{X}$ of $B$ samples $x_1, \cdots, x_B$, the objective is that for every $x$ there exists one and only
one neuron $n_i$ such that $L(x)_i > 0$ and $L(x')_i < 0$ for all $x' \neq x$, where $L(x)_i$ denotes the activation of
the $i$-th neuron for sample $x$. This ensures that neuron $n_i$ allows for perfect reconstruction of $x$ from the
gradients of its weight row, while producing zero gradients for all other samples. Therefore, for every
neuron $n_i$ the desired probability to activate for any sample in the batch is $1/B$. We take the model's weights
and our approximated normalized features to be independently and identically distributed (i.i.d.) random variables
sampled from a normal distribution:
\begin{equation}
    \begin{pmatrix}
        \bf{w}_i\\
        \bf{x}_i
    \end{pmatrix}\sim\mathcal{N}(0,I_M)
\end{equation}
Using this assumption, the probability of a single neuron, with corresponding weight row $w_i$ and bias $b_i$,
activating for a sample $x_i \in \mathbb{R}^{M}$ can be expressed as:
\begin{equation}
    P\bigl(L(x_i)>0\bigr) = P\bigl(w_i^{T}x_i+b_i > 0 \bigr),
\end{equation}
which can be expanded as:
\begin{equation}\label{eq:summErweitert2}
\begin{split}  
     P\bigl(w_{i1}x_{i1}+\ldots +w_{iM}x_{iM}+b_i > 0 \bigr)
\end{split}
\end{equation}
Now the left-hand side is a sum of i.i.d. random variables, each of which has characteristic function:
\begin{equation}\label{eq:charakteristisch2}
\bigl(1+t^2\bigr)^{\nicefrac{1}{2}}
\end{equation}
In turn, the characteristic function of the entire sum is:
\begin{equation}
(1+2it)^{-M/2}=(1-2it/2)^{-M/2}\cdot(1+2it/2)^{-M/2}
\end{equation}
This immediately identifies it as distributed like:
\begin{equation}
    P\bigl(w_i^Tx_i+b_i>0\bigr)=P\bigl(Q_1/2 - Q_2/2\leq b_i\bigr)
\end{equation}
where $Q_1$ and $Q_2$ are independent and both of the chi-square distribution with $M$ degrees of freedom. Note
that the variance of each addend is $V(w_{ij}x_{ij})=1$ (derived by evaluating the second derivative
of~\Cref{eq:charakteristisch2} at 0). We can drop the assumption that our features are normally-distributed
and instead assume that the data has been normalized, as we do in our experiments. Even with this change,
the variance of each addend still evaluates to $1$ as we maintain the assumption that the weights are
independently- and normally-distributed. This enables us to apply the Central Limit Theorem to the
original sum in~\Cref{eq:summErweitert}. As a sum of i.i.d. random variables with existing second moments,
we can make a statement about convergence in distribution:
\begin{equation}
    P\bigl(w_{i1}x_{i1}+\ldots +w_{iM}x_{iM}+b_i > 0 \bigr)\xrightarrow{d}\Phi\biggl(\frac{b_i}{\sqrt{M}}\biggr)
\end{equation}
where $\Phi$ is the cumulative distribution function of a standard normal distribution.
Meaning we can solve for the asymptotically optimal bias $b_*$ by using the inverse cdf or quantile function:
\begin{equation}
b_*=\Phi^{-1}(\frac{1}{B})\cdot \sqrt{M}
\end{equation}

It can be helpful to conceptualize the relationship between this linear layer and a set of samples as a
bipartite random graph. The nodes of one partition class are neurons from the neuron
set  $\smash{\tilde{N}}=\{n_1,\ldots ,n_n\}$, and those in the other partition are from the sample
set $\smash{\tilde{X}}=\{x_1,\ldots ,x_B\}$. A vertex $(n_i, x_j)\in\smash{\tilde{N}}\times\smash{\tilde{X}}$ is
said to exist if $n_i$ is activated by $x_j$. If the malicious actor wants to reconstruct a
sample $x_j$, that sample needs to have a neighbor $n_i$ with degree $\delta(n_i)=1$.

\paragraph{Extraction Metrics} We closely follow the extraction metrics proposed by~\citet{boenisch2023curious}.
Given a linear layer with $N$ output neurons and a batch $\tilde{X}$ of $B$ samples $x_1, \cdots, x_B$, we define
the following metrics:
\begin{enumerate}
    \item \textbf{Active neurons} ($A$): This metric represents the percentage of neurons in
    layer $L$ that activate for at least one of the $B$ samples. Formally, let $N_A$ be the
    number of neurons $n_i$ that satisfy the condition that there exists at least one
    input $x$ in $\tilde{X}$ such that $L(x)_i > 0$, where $L(x)_i$ denotes the activation
    of the $i$-th neuron in $L$ for sample $x$. For the neuron conceptualized as a graph node,
    this condition is equivalent to $\delta(n_i) \geq 1$. The active neurons metric $A$ is therefore
    defined as the ratio of $N_A$ to the total number of neurons $N$:
    \begin{equation}
        A = \frac{N_A}{N}
    \end{equation}
    \item \textbf{Extraction-Precision} ($P$): This metric measures the percentage of neurons that allow for
    the extraction of individual data points. Specifically, let $N_u$ be the number of
    neurons $n_i$ in layer $L$ with unique activations, i.e. those that satisfy the following
    condition: $L(x)_i > 0$ for one input $x$ in $\tilde{X}$, and $L(x')_i < 0$ for all other
    inputs $x' \neq x$. This condition is equivalent to $\delta(n_i)=1$, which effectively counts
    neuron nodes that are leaves of their graphs. The extraction-precision $P$ is defined
    as the following ratio:
    \begin{equation}
        P = \frac{N_u}{N}
    \end{equation}
    \item \textbf{Extraction-Recall} ($R$): The extraction-recall measures the percentage of input
    data points that can be perfectly reconstructed from any gradient row. Let $B_0$ be the
    number of data points that can be extracted with an $l_2$-error of zero, then $R$ is denoted as:
    \begin{equation}\label{eq:Rdef2}
    R=\frac{B_0}{B}
    \end{equation}
\end{enumerate}
If the true probability of a neuron
activating is $1/B$, we can derive the explicit probabilities $p_{A;B}$ and $p_{u;B}$ for a
neuron to be counted as a success in the context of the $A$ and $P$ metrics, respectively.
Since the success of one neuron does not influence the remaining neurons, the entire batch
follows a binomial distribution: $N_A\sim\mathcal{B}(N,p_{A;B})$ and $N_u\sim\mathcal{B}(N,p_{u;B})$.
Consequently, the expected activation share $A$ and the expected extraction-precision $P$ are:
\begin{equation}
    p_{A;B}=\mathbb{E}[A_B]=1-\bigl(\frac{B-1}{B}\bigr)^B
\end{equation}
and
\begin{equation}
    p_{u;B}=\mathbb{E}[P_B]=\bigl(\frac{B-1}{B}\bigr)^{B-1}
\end{equation}

For growing batch sizes, these converge to:
 \begin{equation}
     \lim\limits_{B \to \infty} \mathbb{E}[A_B] = 1-\frac{1}{e}\approx 63.2\%
 \end{equation}
and
  \begin{equation}
     \lim\limits_{B \to \infty} \mathbb{E}[P_B] = \frac{1}{e}\approx 36.8\%.
 \end{equation}
From a graph theory perspective, $R \cdot B = B_0$ denotes the size of the largest neuron subset $V \subset \tilde{N}$,
such that the subgraph induced by the union of $V$ and all neighbors of vertices in $V$ forms a
perfect matching. However in contrast to the previous metrics, the expected extraction recall $R$ exhibits
a crucial difference. Our assumptions state that existence of a particular edge (representing the activation
of a neuron by a data point) is independent of the existence of any other edges. Due to this, and
because neurons do not have edges in common with one another, the event of inclusion of any neuron
in the count for $N_A$ and $N_u$ is also independent of the inclusion of any other neuron therein.
This allows us to assert the success probabilities and that $N_A$ and $N_u$ will
both follow a binomial distribution. This binomial approach breaks down for $R$, however. We can and
will still derive the non-zero probability for one specific data point to be included
in the count (for $N_A$ and $N_u$ we were counting neurons, for $B_0$ we count data points).
This value will depend on the size of both partition classes of the graph, not just the one being counted.
Let $B$ be the number of samples in our batch, and $N$ be the number of neurons in our linear layer. Additionally,
we assume that we have achieved the optimal probability of activation of $1/B$ for every neuron-sample pair. Given
a single neuron $n_i$ and a single sample $x$, the probability that this neuron activates only for $x$ and not for
all other samples $x' \neq x$ is
\begin{equation}
    \frac{1}{B}\left(\frac{B-1}{B}\right)^{B-1},
\end{equation}
which we also refer to as the probability that $n_i$ isolates $x$. The complement of this event is the probability
that the neuron does \textbf{not} isolate $x$:
\begin{equation}
    1 - \frac{1}{B}\left(\frac{B-1}{B}\right)^{B-1}.
\end{equation}
Raising this result to the power of $N$ yields the probability that all neurons do \textbf{not} isolate $x$. Finally,
the complement of this event is the probability that at least one neuron isolates $x$, which in turn results in $x$
being reconstructed.
Namely, the expected share of data points that the malicious actor can perfectly reconstruct is:
\begin{equation}
    p_{R;B;N} = \mathbb{E}[R] = 1 - \left(1 - \frac{1}{B}\left(\frac{B-1}{B}\right)^{B-1}\right)^N
\end{equation}
To check that this does not yield the exact
distribution of $R$ and that the binomial approach is no longer relevant, consider a graph with more data
points than neurons, i.e.\ $B>N$. It is clear that $P(R=100\%) = 0\neq (p_{R;B;N})^{\textsc{B}}$, since there
are not enough neurons to cover each data point.

\section{Additional Experimental Results}
\label{appendixResults}

\begin{table}[t]
\centering
\caption{Comparing the percentage of perfectly reconstructed images ($R$, see~\Cref{eq:R}) taken from
    the CIFAR-10 dataset, with the results reported by \citet{boenisch2023curious} using
    their \textit{trap weights (TWs)} approach, across various combinations of neuron counts $N$ and batch sizes $B$.
    The listed values represent the mean across 10 random initializations, with each initialization
    evaluated on 10 random batches. The error margins indicate the 95\% confidence interval.}\label{table:imageResults2}
\begin{tabular}{@{}cccc@{}}
\toprule
                 \multicolumn{4}{c}{\textbf{CIFAR-10}}                     \\ 
\textbf{(N, B)} & \textit{TWs} & \textit{QBI}    & \textit{PAIRS}           \\ \midrule
(200, 20)       & 69.5         & 75.7 $\pm$ 1.85 & \textbf{77.1 $\pm$ 1.19} \\
(200, 50)       & 45.2         & 46.5 $\pm$ 1.06 & \textbf{48.4 $\pm$ 0.43} \\
(200, 100)      & 26.9         & 28.4 $\pm$ 0.60 & \textbf{31.5 $\pm$ 0.73} \\
(200, 200)      & 9.60         & 15.8 $\pm$ 0.49 & \textbf{18.6 $\pm$ 0.32} \\
(500, 20)       & 87.0         & 87.6 $\pm$ 1.18 & \textbf{87.8 $\pm$ 1.36} \\
(500, 50)       & 61.4         & 63.8 $\pm$ 1.09 & \textbf{67.3 $\pm$ 1.28} \\
(500, 100)      & 42.2         & 45.1 $\pm$ 0.74 & \textbf{48.1 $\pm$ 0.80} \\
(500, 200)      & 17.7         & 28.4 $\pm$ 0.43 & \textbf{32.6 $\pm$ 0.60} \\
(1000, 20)      & 91.5         & 91.3 $\pm$ 1.40 & \textbf{91.8 $\pm$ 1.49} \\
(1000, 50)      & 72.4         & 74.3 $\pm$ 0.93 & \textbf{77.7 $\pm$ 1.17} \\
(1000, 100)     & 55.6         & 57.2 $\pm$ 0.76 & \textbf{59.4 $\pm$ 0.52} \\
(1000, 200)     & 25.6         & 39.2 $\pm$ 0.49 & \textbf{42.6 $\pm$ 0.60} \\ \bottomrule
\end{tabular}
\end{table}

\begin{table}[t]
\caption{Comparing $A$ and $P$ obtained on the IMDB text dataset, (see~\Cref{eq:defA,eq:defP}) with the results reported
by \citet{boenisch2023curious}, across various batch sizes $B$ using a layer size of 1000. The corresponding $R$
    values measuring the extraction success are listed in \Cref{table:textResults}. Results are averaged over
    10 random initializations, each evaluated on 10 random batches.}
\label{table:APText}
\centering
\begin{tabular}{@{}lllllll@{}}
\toprule
\textbf{}  & \multicolumn{3}{c}{A}                & \multicolumn{3}{c}{P}                \\
\textbf{B} & \textit{TWs} & \textit{QBI} & \textit{PAIRS} & \textit{TWs} & \textit{QBI} & \textit{PAIRS} \\ \midrule
20         & 51.9     & 58.7       & 59.2         & 61.0     & 33.3       & 33.6         \\
50         & 77.6     & 57.1       & 57.0           & 37.6     & 32.6       & 32.5         \\
100        & 91.0     & 55.1       & 55.5         & 19.2     & 31.8       & 31.5         \\
200        & 97.8     & 53.4       & 53.8         & 0.07     & 30.6       & 30.7         \\ \bottomrule
\end{tabular}
\end{table}

\begin{table}[t]
\centering

\caption{Comparing $A$ and $P$ (see~\Cref{eq:defA,eq:defP}) obtained on the ImageNet dataset with the results reported by \citet{boenisch2023curious},
    across various batch sizes $B$ and layer sizes $N$. The corresponding $R$ values measuring the extraction
    success are listed in \Cref{table:imageResults}. Results are averaged over 10 random initializations, each
    evaluated on 10 random batches. }\label{table:APImage}

\begin{tabular}{@{}ccccccc@{}}
\toprule
\multicolumn{1}{l}{} & \multicolumn{3}{c}{A}                & \multicolumn{3}{c}{P}                \\
\textbf{(N, B)}        & TWs  & \textit{QBI} & \textit{PRS} & TWs  & \textit{QBI} & \textit{PRS} \\ \midrule
(200, 20)            & 0.09 & 75.6         & 72.6           & 94.8 & 31.4         & 33.9           \\
(200, 50)            & 38.1 & 80.8         & 76.2           & 76.3 & 25.1         & 29.7           \\
(200, 100)           & 65.3 & 84.6         & 79.0           & 50.0 & 21.4         & 27.8           \\
(200, 200)           & 88.6 & 86.5         & 80.6           & 23.3 & 18.8         & 26.3           \\
(500, 20)            & 0.09 & 75.8         & 72.7           & 93.9 & 31.4         & 33.0           \\
(500, 50)            & 38.7 & 81.4         & 76.2           & 76.7 & 26.2         & 30.2           \\
(500, 100)           & 64.6 & 84.6         & 78.6           & 50.8 & 21.6         & 27.7           \\
(500, 200)           & 89.2 & 87.1         & 80.4           & 24.0 & 18.8         & 26.3           \\
(1000, 20)           & 10.2 & 75.5         & 73.4           & 94.2 & 31.4         & 33.1           \\
(1000, 50)           & 38.8 & 80.9         & 76.4           & 77.0 & 26.1         & 30.1           \\
(1000, 100)          & 65.5 & 84.4         & 79.3           & 51.4 & 22.0         & 28.0           \\
(1000, 200)          & 89.2 & 87.3         & 81.5           & 23.8 & 18.8         & 26.0           \\ \bottomrule
\end{tabular}
\end{table}

\begin{figure*}[t]
  \includegraphics[width=\textwidth]{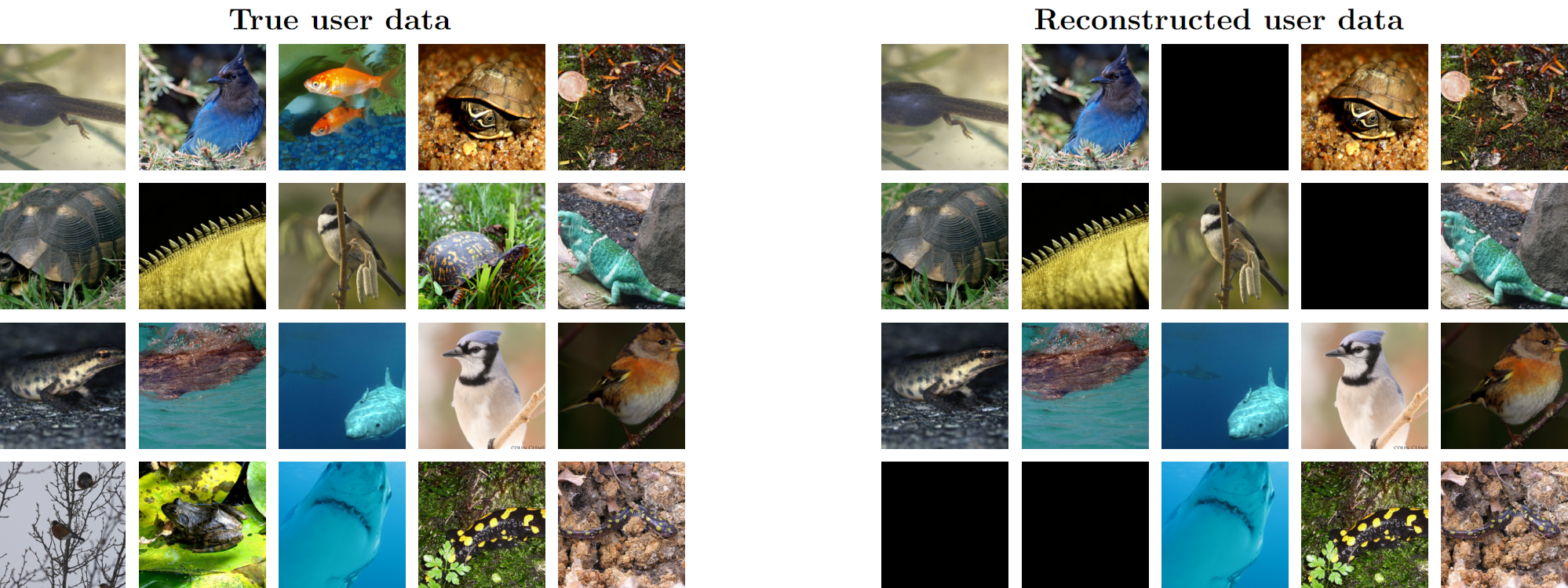}
    \caption{True user data (left), a batch of 20 images from the ImageNet dataset and reconstructed user data (right),
        using a linear layer of size 200 that was maliciously initialized with our QBI approach. Fully black images
        denote data points that could not be recovered. Despite the small layer size, in this particular setting,
        our method achieves perfect reconstruction of around 82.5\% of the original data points, on average.}
  \label{fig:user_data}
  
\end{figure*}

\begin{table}[t]
\centering
\caption{Comparing the predicted values $A_{pr}$, $P_{pr}$ and $R_{pr}$, obtained via~\Cref{eq:optimalA,eq:optimalP,eq:R},
    to $A_{ex}$, $P_{ex}$ and $R_{ex}$ observed experimentally when using a synthetic, fully random dataset.
    All numbers are averaged over 300 random initalizations, tested with 10 batches of random data each.
    The experimental setup was idential to that when evaluating the extraction on the CIFAR-10 dataset using QBI,
    replacing the image data with normal random noise of shape $3\times32\times32$.}
\label{table:synthetic}
\begin{tabular}{@{}cllllll@{}}
\toprule
\textbf{(N, B)} & \multicolumn{1}{c}{\textit{$A_{pr}$}} & \multicolumn{1}{c}{\textit{$A_{ex}$}} & \multicolumn{1}{c}{\textit{$P_{pr}$}} & \multicolumn{1}{c}{\textit{$P_{ex}$}} & \multicolumn{1}{c}{\textit{$R_{pr}$}} & \multicolumn{1}{c}{\textit{$R_{ex}$}} \\ \midrule
(200, 20)     & 64.2                                 & 64.1                                & 37.7                                 & 37.5                                & 97.8                                 & 97.7                                \\
(200, 50)     & 63.6                                & 64.2                                & 37.2                                 & 37.3                                & 77.5                                 & 77.1                                \\
(200, 100)    & 63.4                                 & 63.0                                & 37.0                                 & 36.7                                & 52.3                                 & 52.1                                \\
(200, 200)    & 63.3                                 & 63.1                                & 36.9                                 & 36.5                                & 30.9                                 & 30.7                                \\
(500, 20)     & 64.2                                 & 64.2                                & 37.7                                 & 38.0                                & 100                                  & 100                                 \\
(500, 50)     & 63.6                                 & 63.4                                & 37.2                                 & 37.0                                & 97.6                                 & 97.5                                \\
(500, 100)    & 63.4                                 & 63.3                                & 37.0                                 & 36.8                                & 84.3                                 & 83.9                                \\
(500, 200)    & 63.3                                 & 63.1                                & 36.9                                 & 36.5                                & 60.3                                 & 59.8                                \\
(1000, 20)    & 64.2                                 & 64.2                                & 37.7                                 & 37.6                                & 100                                  & 100                                 \\
(1000, 50)    & 63.6                                 & 63.6                                & 37.2                                 & 37.0                                & 99.9                                 & 100                                 \\
(1000, 100)   & 63.4                                 & 63.2                                & 37.0                                 & 36.8                                & 97.5                                 & 97.1                                \\
(1000, 200)   & 63.3                                 & 63.4                                & 36.9                                 & 36.8                                & 84.2                                 & 83.6                                \\ \bottomrule
\end{tabular}
\end{table}

\begin{table}[t]
\caption{Comparing extraction recall $R$ on image data achieved using QBI in combination with regular data
normalization (DN), batch normalization (BN) and layer normalization (LN) across ImageNet and CIFAR-10.
The values represent the mean across 10 random initializations, with each initialization
evaluated on 10 random batches. Error margins indicating the 95\% confidence interval ranged from $\pm 0.2$ to $\pm 1.57$ and
are omitted for readability. (*) For layer normalization, extraction recall is defined as the percentage of samples
that are isolated by a single neuron, see \Cref{app:layernorm} for further explanation.}
\label{table:norms}
\centering
\begin{tabular}{@{}ccccccc@{}}
\toprule
                & \multicolumn{3}{c}{ImageNet}                                                     & \multicolumn{3}{c}{CIFAR-10}                                                     \\
\textbf{(N, B)} & \multicolumn{1}{l}{\textit{DN}} & \textit{BN} & \textit{LN*} & \multicolumn{1}{l}{\textit{DN}} & \textit{BN} & \textit{LN*} \\ \midrule
(200, 20)       & 82.5                                  & 81.3               & \textbf{95.5}       & 75.7                                  & 77.7               & \textbf{94.8}       \\
(200, 50)       & 52.0                                  & 51.9               & \textbf{70.2}       & 46.5                                  & 47.6               & \textbf{67.7}       \\
(200, 100)      & 29.0                                  & 32.5               & \textbf{44.2}       & 28.4                                  & 28.9               & \textbf{39.1}       \\
(200, 200)      & 15.1                                  & 18.7               & \textbf{24.2}       & 15.8                                  & 16.1               & \textbf{21.7}       \\
(500, 20)       & 93.6                                  & 91.4               & \textbf{99.9}       & 87.6                                  & 87.6               & \textbf{99.5}       \\
(500, 50)       & 73.5                                  & 71.3               & \textbf{94.4}       & 63.8                                  & 63.9               & \textbf{90.4}       \\
(500, 100)      & 49.0                                  & 50.0               & \textbf{75.0}         & 45.1                                  & 44.8               & \textbf{70.2}       \\
(500, 200)      & 29.2                                  & 32.6               & \textbf{49.3}       & 28.4                                  & 28.0               & \textbf{44.0}         \\
(1000, 20)      & 96.6                                  & 94.9               & \textbf{100}        & 91.3                                  & 91.6               & \textbf{99.9}       \\
(1000, 50)      & 84.3                                  & 81.4               & \textbf{99.6}       & 74.3                                  & 74.5               & \textbf{98.8}       \\
(1000, 100)     & 64.8                                  & 63.4               & \textbf{92.4}       & 57.2                                  & 56.2               & \textbf{89.5}       \\
(1000, 200)     & 42.7                                  & 44.1               & \textbf{72.4}       & 39.2                                  & 38.7               & \textbf{66.8}       \\ \bottomrule
\end{tabular}
\end{table}

\subsection{AGGP}
\label{appendix:aggp}

\begin{figure}[t]
    \centering
    \includegraphics[width=1\columnwidth]{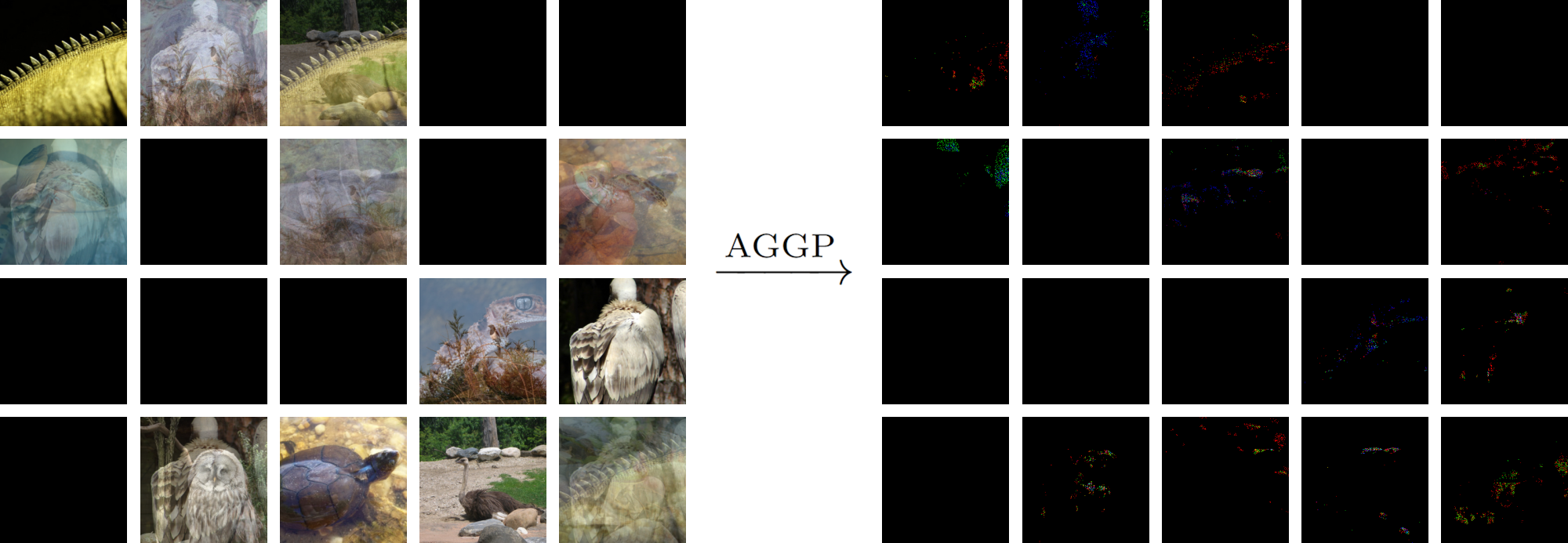}
    \caption{Visualization of the active data leakage of the first 20 neurons of a linear layer of size 200 (left),
        that was maliciously initialized using QBI, and the impact of AGGP (right). The artificially induced sparsity
        leads to aggressive gradient pruning across the entire layer.}
    \label{fig:active_leak}
\end{figure}
\begin{figure}[t]
  \centering
  \includegraphics[width=\columnwidth]{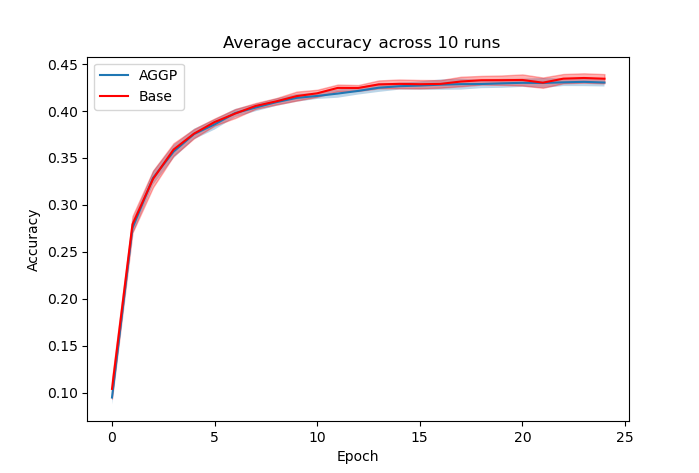}
  \caption{Performance of a benign CNN-based image model (\Cref{table:imageModel}) on the CIFAR-10 dataset, using a
  batch size of 64. Comparing the unmodified version (Base) to one protected using AGGP.  The experiment used the Adam
  optimizer \cite{kingma2014adam} with a learning rate of 1e-5 and optimized the cross-entropy loss. Results are
  averaged across 10 runs using different seeds. The shaded regions correspond to the 95\% confidence interval.}
  \label{fig:aggpEval}
\end{figure}

We evaluated the impact of AGGP on a CNN-based architecture (see~\Cref{table:imageModel}), which could, in theory,
be maliciously initialized by changing parameter values without modifying the architecture. Since model performance
is not a concern when the central entity is malicious or compromised, the impact is assessed on a benign network.
The model's validation accuracy on the CIFAR-10 dataset was evaluated across 10 random initializations, recorded
across 25 epochs, using a batch size of 64. The results of these preliminary experiments, presented
in~\Cref{fig:aggpEval}, show no significant impact on training performance.

\subsection{Extraction under Batch Normalization}
\label{app:batchnorm}
When a normalization layer, such as a batch normalization layer, precedes the maliciously initialized linear layer,
the reconstruction process must be adapted accordingly. This section will specifically focus on
the \textit{BatchNorm2d} layer as it is implemented in PyTorch.
Batch normalization uses the following equation to perform normalization:

\begin{equation}
    y = \frac{x-\text{E}[x]}{\sqrt{\text{Var}[x]+\epsilon}}*\gamma+\beta
    \label{eq:bn}
\end{equation}

where $x$ is the input, $\gamma$ and $\beta$ are learnable parameters with default values $1$ and $0$ respectively,
and $\epsilon$ is a small constant added to the denominator for numerical stability, with default value $10^{-5}$.
The mean and standard-deviation are calculated per-dimension over the mini-batches. Additionally \textit{BatchNorm2d}
keeps running estimates of its computed mean $\hat{\mu}$ and variance $\smash{\hat{V}}$, which are both updated using
a \textit{momentum} of $0.1$ using the following equation:
\begin{equation}
    \hat{x}_{new} = (1-\textit{momentum}) \cdot \hat{x} + \textit{momentum} \cdot x_t
\end{equation}
where $\hat{x}$ is the estimated statistic and $x_{t}$ is the new observed value. As extraction is only done on
the initial training step after model initialization, $\hat{\mu}$ and $\hat{V}$ will be set to their default values
0 and 1 respectively. Hence, the values that we obtain after the first training step are:
\begin{equation}
    \hat{\mu}_{new} = \textit{momentum} \cdot \mu_t
\end{equation} 
and
\begin{equation}
    \hat{V}_{new} = (1-\textit{momentum}) + \textit{momentum} \cdot V_t
\end{equation}
allowing us to simply rearrange the equations to solve for the batch statistics:
\begin{equation}
    \hat{\mu}_{t} = \frac{\hat{\mu}_{new}}{\textit{momentum}}
\end{equation} 
\begin{equation}
    \hat{V}_{t} = \frac{\hat{V}_{new} - (1 - \textit{momentum})}{\textit{momentum}}
\end{equation}

Since the learnable parameters $\gamma$ and $\beta$ are intialized to 1 and 0 respectively, they have no effect during the
first training step and can be ignored. Finally, given the extracted sample $y$ from the gradients, normalization
can be reversed to obtain the original sample $x$ by rearranging equation \Cref{eq:bn} as:
\begin{equation}
    x =\hat{\mu}_{t} +  y  \sqrt{\hat{V}_t
     +\epsilon } 
\end{equation}

\subsection{Extraction under Layer Normalization}
\label{app:layernorm}
\begin{figure}[t]
    \centering
    \includegraphics[width=\columnwidth]{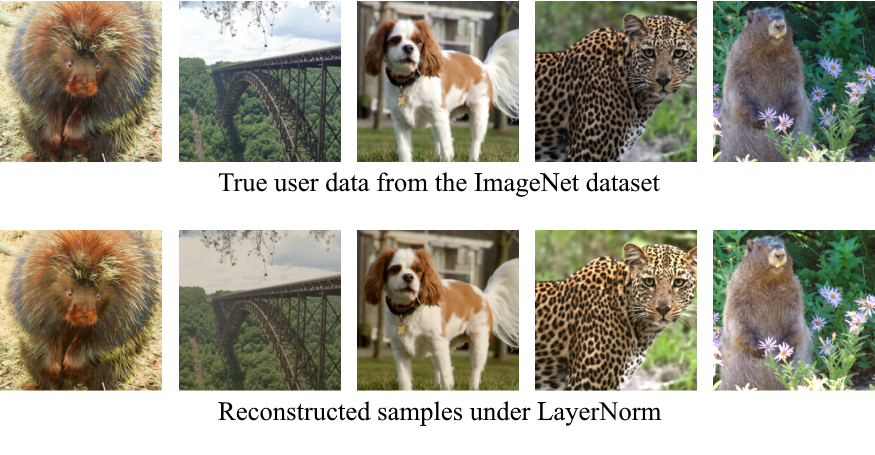}
    \caption{A mini-batch of true user data from the ImageNet dataset, compared to the images that could be
    reconstructed when the QBI-initialized linear layer was preceded by a \textit{LayerNorm} layer. Normalization
    was reversed using the publicly available ImageNet normalization parameters. The reversal of normalization with
    imperfect parameters introduces a slight shift to the images color and brightness, however detail and structure
    are preserved, leading to a high structural similarity index (SSIM), ranging from $0.82$ to $0.96$ for these
    samples.}
    \label{fig:layernorm}
\end{figure}

 This section will specifically focus on the \textit{LayerNorm} layer as it is implemented in PyTorch.
Layer normalization uses the same equation as batch normalization (see \Cref{eq:bn}) with the difference, that the
mean and standard-deviation are calculated over the last $D$ dimensions of every sample, where $D$ is the dimension
of the \textit{normalized\_shape}. In our case, \textit{normalized\_shape} is a 1-dimensional vector representing the
flattened input image. Since each sample is normalized with respect to its own mean and standard deviation, perfect
reconstruction, defined by an L2 loss of zero, is not possible, as the central entity does not have knowledge of these
per-sample normalization parameters which are typically not shared or cached. However, normalization can be reversed
by substituting with commonly available normalization parameters of the target domain, as an approximation for the true
sample statistics. Figure \ref{fig:layernorm} displays a mini-batch of true user data compared to the images that
could be reconstructed when the QBI-initialized linear layer was preceded by a \textit{LayerNorm} layer, where
normalization was reversed using the publicly available ImageNet normalization parameters. It is evident, that
while the reversal of normalization with approximated parameters introduces a slight shift to the images color
and brightness, detail and structure are preserved. When a \textit{LayerNorm} layer precedes a linear layer that
was maliciously initialized with QBI, reconstruction success, as measured by samples that are isolated by a single
neuron, is greatly increased, as per-sample normalization is more effective at bringing input features closer to
being normally distributed, compared to regular data normalization or batch normalization.

\subsection{AGGP Hyperparameters}
\label{app:aggphyperparams}

\begin{figure}[t]
  \centering
  \includegraphics[width=1\columnwidth]{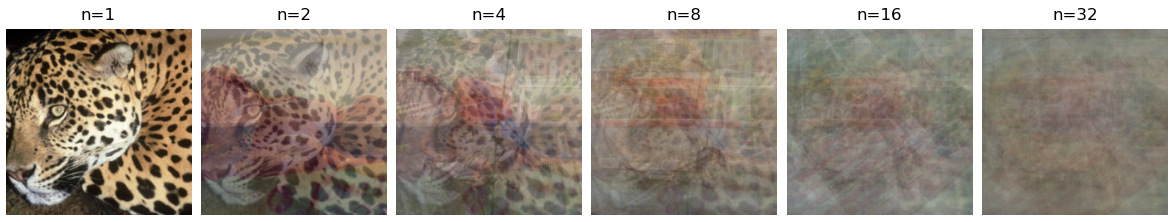}
  \caption{Effect of averaging an increasing number of images $n$ from the ImageNet dataset, on the level of obfuscation.}
  \label{fig:overlay}
\end{figure}

\begin{figure}[t]
  \centering
  \includegraphics[width=\columnwidth]{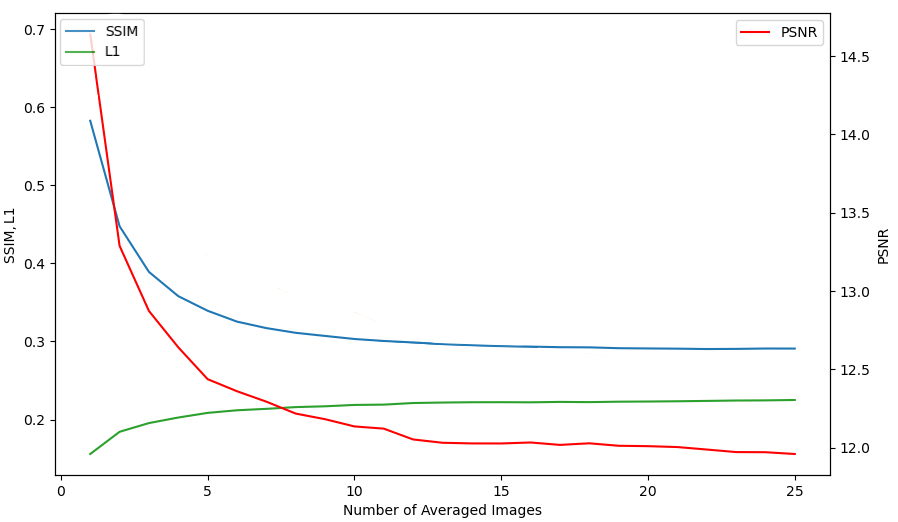}
  \caption{}
  \label{fig:psnr}
  \caption{Average similarity metrics (SSIM, L1 Distance, PSNR) between an original image from the ImageNet
  dataset and its average with 1 to 25 additional images, repeated 100 times. PSNR and SSIM drop significantly
  up to N=15, while L1 Distance increases.}
\end{figure}

To determine suitable hyperparameters for AGGP applied to ImageNet data, we investigated the level of obfuscation
achieved when multiple images are averaged together in a gradient row.
\Cref{fig:overlay} displays an example of this effect, where the average is computed over an increasing number of
images $n$. To quantify this effect more objectively, we conducted a systematic evaluation. We randomly selected
one image and averaged it with an increasing number of additional images (from 1 to 25), and then computed three
commonly used metrics to quantify the similarity between the original image and the averaged image: the Structural
Similarity Index Measure (SSIM, \cite{ssim}), L1 Distance, and Peak Signal-to-Noise Ratio (PSNR). By repeating this
process 100 times and averaging the metrics across all experiments, we obtained the results shown in \Cref{fig:psnr}.
The results clearly show that PSNR and SSIM decrease sharply in the range $[0,15]$, while the L1 distance increases.
After that, the values appear to converge slowly with little to no movement, which is why we decided to select 16 as
our cut-off value $c$. As evident from \Cref{fig:overlay} and \Cref{fig:psnr}, low activation counts provide little
obscurity, while higher activation counts require minimal pruning to obscure data, which is why we set the bounds
for $p_\text{keep}$ to $p_l=0.01$ and $p_u=0.95$.

\section{Experiment details}
\label{section:implDetails}
All results that we report are averaged across 10 runs, each evaluated on 10 batches of unseen test data.
For each individual run, the weights are randomly initialized, and the train / test split is shuffled randomly.
In the case of malicious model initialization and data extraction, train images refer to those used by the PAIRS
algorithm to tune the model's parameters, while test images refer to those used to evaluate the reconstruction
percentage. For each (N,B) setting, the standard error across these 10 runs is calculated, and multiplied by
$1.96$ to determine the 95\% CI. For every batch, a single forward pass is performed, and the metrics $A$, $P$
and $R$ (see~\Cref{eq:defA,eq:defP,eq:Rdef}) are obtained. Since during evaluation, we have access to the internals of the
model, we don't have to examine the gradients to determine the data leakage. Rather we directly determine which
samples will be leaked by observing the activation patterns in the maliciously initialized layer during the
forward pass. Furthermore this direct access allows us to feed the input directly to our maliciously initialized
layer, circumventing compute heavy convolutional layers that were initialized as identity functions.

\subsection{Datasets}
\label{section:datasets}
We used three datasets to evaluate our method: The ImageNet-1k \cite{imagenet} validation set (6GB, 50k images), the CIFAR-10 \cite{cifar} dataset and
the IMDB \cite{imdb} binary sentiment classification dataset. All these datasets can be used for non-commercial research purposes.

\subsection{Compute Resources}
\label{section:compute}
All experiments were run on an RTX 2060 Super with 8GB VRAM. Data-extraction runs in near real-time,
scaling linearly with batch size and data dimensionality, as it is simply done in one step, by dividing the
weight gradients by the bias gradients. Time to maliciously initialize the model varied across methods: QBI
required less than 1 second in all settings, while PAIRS initialization times ranged from 10 seconds for
CIFAR-10 (N=200, B=20) to 12 minutes for ImageNet (N=1000, B=200). For even larger layer sizes N, initialization
time for PAIRS will scale linearly. Obtaining all results, for both QBI and PAIRS, across all possible combinations
of N and B and all three datasets, averaged across 10 runs with random seeds for model intialization and
train / test split, required approximately 8 hours of compute. Evaluation of AGGP's impact on training
performance across 20 runs (10 Base, 10 protected with AGGP) required about 2 hours of training time.
Experiments evaluating QBI on synthetic data (see~\Cref{table:synthetic}) took about 1 hour, averaging
over 300 random runs per (N, B) setting.

\subsection{Image Models}
\label{section:imageModels}
\Cref{table:imageModel} outlines the implementation of the model used for image data extraction.
As \citet{boenisch2023curious} explain in their Appendix B, convolutional layers can be modified
to pass the input to the next layer, effectively acting as an identity function. This can be achieved
through various methods. \Cref{algo:cnnIdentity} provides a minimal working example using a 2D convolutional
layer, suitable for RGB images. To preserve the image shape, we employ a kernel size of 3, padding of 1, and
stride of 1. Since we have three channels to transmit, we initialize three filters, each acting as the identity
function for its respective channel. This is achieved by setting the weight values of the $i$-th filter to zero
and then setting the center value of its weight matrix for the $i$-th channel to one. Additionally, randomly
initialized filters could be added to obscure the modifications made to the model.

\begin{algorithm}
\caption{Conv2D Identity Initialization Example}
\begin{algorithmic}[1]
  \State $num\_channels \gets 3$
  \State $conv2d \gets \Call{Conv2d}{in=3, out=3, k=3, s=1, p=1}$
   
  \For{$i \gets 0$ to $num\_channels$}
    \State $conv2d.weight.data[i, :, :, :] \gets 0$
    \State $conv2d.weight.data[i, i, 1, 1] \gets 1$
  \EndFor
\end{algorithmic}
\label{algo:cnnIdentity}
\end{algorithm}

\label{appendix:imp_details}

\begin{table}[H]
\centering
\caption{Architecture of models used in the experiments on image data. f: number of filters, k: kernel size,
    s: stride, p: padding act: activation function, n: number of neurons. The size of the second to last layer
    was varied across experiments.}
\label{table:imageModel}
\begin{tabular}{@{}c@{}}
\toprule
CNN Architecture                \\ \midrule
Conv(f=128, k=(3, 3), s=1, p=1) \\
Conv(f=256, k=(3, 3), s=1, p=1) \\
Conv(f=3, k=(3, 3), s=1, p=1)   \\
Flatten                         \\
$<$Optional$>$ BatchNorm / LayerNorm \\
Dense(n=1000, act=ReLU)          \\
Dense(n=\#classes, act=None)    \\ \bottomrule
\end{tabular}
\end{table}

\begin{table}[H]
\centering
\caption{Architecture of models used in the experiments on the IMDB dataset. feat: vocabulary size, dim:
embedding size, act: activation function, n: number of
neurons.}
\label{table:textModel}
\begin{tabular}{@{}c@{}}
\toprule
IMDB-Model Architecture          \\ \midrule
Embedding(feat=10\_000, dim=250) \\
$<$Optional$>$ BatchNorm / LayerNorm \\
Dense(n=1000, act=ReLU)          \\
Dense(n=1, act=None)             \\ \bottomrule
\end{tabular}
\end{table}

\end{document}